\pdfoutput=1
\documentclass[11pt]{article}

\usepackage[final]{acl}

\usepackage{times}
\usepackage{latexsym}

\usepackage[T1]{fontenc}
\usepackage[utf8]{inputenc}

\usepackage{microtype}

\usepackage{inconsolata}
\usepackage{graphicx}
\usepackage{subcaption}

\title{Automated Answer Validation using Text Similarity}

\author{
    Balaji Ganesan  \\
    IBM Research India \\
    bganesa1@in.ibm.com \\\And
      Arjun Ravikumar  \\
      Independent Researcher \\
  ark.arjun97@gmail.com \\\And
      Lakshay Piplani   \\
           Independent Researcher \\
  Lakshay.piplani94@gmail.com \\\AND
      Rini Bhaumik   \\
           Independent Researcher \\
 rini.bhaumik@gmail.com \\\And
      Dhivya Padmanaban   \\
           Independent Researcher \\
  dhivyathamil@gmail.com \\\And
  Shwetha Narasimhamurthy \\
       Independent Researcher \\
  shwethahybris@gmail.com \\\AND
    Chetan Adhikary  \\
    Tata Consultancy Services \\
  chetanadhikary@gmail.com \\\And
    Subhash Deshapogu  \\
         Independent Researcher \\
  subhashdeshapogu@gmail.com \\
}

\begin{document}
\maketitle
\begin{abstract}
Automated answer validation can help improve learning outcomes by providing appropriate feedback to learners, and by making question answering systems and online learning solutions more widely available. There have been some works in science question answering which show that information retrieval methods outperform neural methods, especially in the multiple choice version of this problem. We implement Siamese neural network models and produce a generalised solution to this problem. We compare our supervised model with other text similarity based solutions.
\end{abstract}

\section{Introduction}
Reading comprehension based question answering, where a passage is given and models have to answer questions based on the passage, has been studied extensively in the past decade. The science question answering problem is relatively harder given the it's a narrower domain. In particular, science question answering in the presence of distractors (known generally as multiple choice answers) remains an active area of research.

SciQ dataset introduced by \cite{Welbl2017CrowdsourcingMC} has questions, multiple choices, correct answer, and a supporting passage. Text similarity solutions are a reasonable choice to solve this problem. Comparing the answers to ground truth in closed-book QA is the most generalized version of the problem. Having multiple options makes the problem slightly harder since distractors can have lots of words in common with the correct answer.

The particular problem that we want to address in this work is \textit{automated answer validation}. Given the question and the answer provided by students, we need to evaluate if the answer is correct, using the question and the supported text. We need to generate a pair between the student answer and the correct answers, where the correct answers are determined from the supporting text. 

\begin{figure}
  \centering
  \includegraphics[height=8cm, width=0.75\columnwidth]{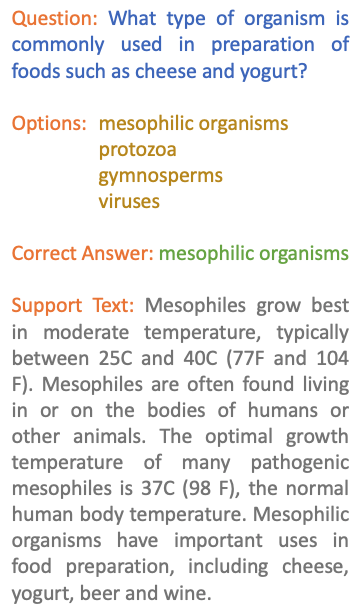}
  \caption{An example question, options, answer and support text from sciq dataset. Our task is to validate the user answer using the support text.}
  \label{fig:sciq_example}
\end{figure}

Consider the example in the SciQ dataset in Figure \ref{fig:sciq_example}. Given the question and the supporting text, we can determine the correct answer is mesophilic organisms. Now we can validate whether the student answer is correct based on text similarity. This is a relatively simple example, where the phrase meosphilic organisms occurs verbatim in the supporting text, but this may not always be the case.

A typical information retrieval based solution to this problem will search the multiple options in the supporting passage. We want to investigate if a neural network model can perform well on this task. Further, if possible, we would like to compare such a supervised model with large language model based solutions. We'll describe our solution in more detail in Section \ref{sec:solution}.

The automated answer validation system for science question answering will benefit educators and students in science-related subjects. It will streamline the assessment process, reduce manual effort, and ensure consistent evaluations, leading to improved learning outcomes in the science domain.

Improving learning outcomes in science education was also the motivation for the SciQ project at Allen Institute for AI. In the Indian context, solving this problem could help reduce the disparities in learning opportunities among different income levels and location of the students. In particular, our solution could help create a scalable platform for students appearing in competitive exams to take mock tests without having to pay for such services in coaching centers.

\section{Related Work}
Approaches to multiple-choice science exam question answering (QA) differ in terms of their reasoning architectures and training processes. \cite{clark2013study} present a collection of sub-issues and methods for solving them. \cite{li2015answering} approach assesses how well a scenario, built from the question and augmented with additional knowledge base (KB) data, holds together. On the other hand, \cite{sachan2016science} develop a model focused on entailment that uses a max-margin ranker to find the right answer based on background knowledge. Models that employ probabilistic reasoning include techniques such as Markov logic networks \cite{khot2015exploring} and an integer linear program-guided approach that constructs proof chains using organized knowledge \cite{khashabi2016question}. The Aristo ensemble \cite{clark2016combining} combines various reasoning approaches with basic statistical methods that rely on lexical relationships and information retrieval (IR). These simpler methods alone offer surprisingly effective starting points. There's been limited effort to apply neural networks to this challenge, mainly due to the scarcity of training data. This paper aims to tackle that problem by creating a substantially larger dataset than what existed before, and we showcase the outcomes of tests conducted using cutting-edge reading comprehension methodologies on our new datasets.

Siamese networks were introduced by \cite{bromley1993signature} who used the method for the task of determining whether two input signatures originated from the same person. Siamese Networks feature two or more identical subnetworks that accept distinct inputs but share the same parameters and weights. These subnetworks are then joined at some point, and the network learns a similarity function over its inputs. The architecture is particularly useful for tasks that involve comparing two different but similar pieces of data, such as in the case of signature or face verification.

Over the years, Siamese Networks have been adapted and extended for various tasks beyond signature verification, including face recognition, object tracking, and more. They have become a foundational architecture in the area of learning similarity metrics and are widely used in various applications today. \cite{singh2019siamese} describes the implementation of siamese network models for text and image similarity.

\cite{viji2022hybrid} describe an hybrid approach that combines BERT extraction with deep Siamese Bi--LSTM model for text similarity. \cite{malkiel2022interpreting} focus on interpretability of BERT based text similarity. \cite{gollapalli2022qsts} proposed using a text similarity measure for question generation.

\cite{li2023using} propose using text similarity for question answering in a particular field of medical domain, namely rheumatoid arthritis question-answering system. \cite{vallejo2022evaluating} discuss the limitations of Pearson Correlation for validating text similarity metrics.

\section{Our approach}
\label{sec:solution}

\begin{figure*}
  \includegraphics[width=\textwidth]{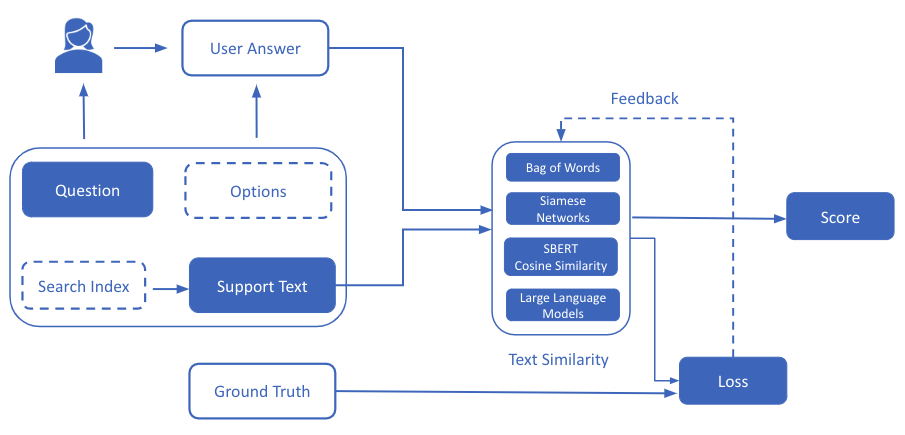}
  \caption{Automated Answer Validation using Text Similarity}
  \label{fig:aae_solution}
\end{figure*}

Our solution for automated answer validation is as shown in Figure \ref{fig:aae_solution}. Given a question, answer, multiple options and supporting text, we use different text similarity methods to predict the correct answer. This answer prediction can be evaluated using ground truth (correct answer) available in the dataset.

We implement a Siamese text similarity model and compare with other methods like few shot inference and retrieval augmented generation using large language models.

\begin{itemize}
    \item Implementing a robust Siamese text similarity model for answer validation in the science domain.
    \item Preprocessing and structuring the dataset from SciQ dataset to create meaningful pairs of student responses and correct answers.
    \item Handling varying lengths of text data during model training and inference.
    \item Addressing potential semantic ambiguities and domain-specific challenges in science questions and responses.
\end{itemize}

After implementing the above model, we compare our solution with an unsupervised method, where SBERT with cosine similarity provides the text similarity score. We also discuss how our solution compares with Large Language Models in Section \ref{sec:extension}. While using LLMs, there are couple of approaches that we could try. Retrieval augmented generation is closer to the information retrieval based solutions in this problem. We can ask the LLM to predict if the selected answer is correct, given the question and the supporting text. As a separate exercise, we could also evaluate LLMs on closed book question answering. In the RAG approach, we can try to extract a relevant portion of the supporting text, to help the LLM make the correct prediction.

In another approach, we can use few shot prompting, where we give the model few examples of the question, multiple choices and correct answers. An instruction fine tuned model like flan-T5 or GPT3.5 could use the samples to validate the answers. We discuss some of these in Section \ref{sec:extension} but do not implement in this work.

\subsection{Methodology}
\begin{itemize}
    \item Combine the question and supporting text to create meaningful context for answer validation in the science domain. Structure the data to form pairs of student responses and correct answers, along with corresponding labels (0 for incorrect and 1 for correct).
    \item Implement a Siamese neural network architecture using NLP libraries and deep learning frameworks. Train the model on the pairs of student responses and correct answers to learn the underlying text representations
    \item Develop an interface for educators to input student responses and query the Siamese text similarity model for validation in the science domain.
    \item Fine-tune the model and update the answer validation system based on evaluation results.
\end{itemize}

\section{Siamese Neural Networks}

\begin{figure*}
  \centering
  \includegraphics[width=0.8\textwidth]{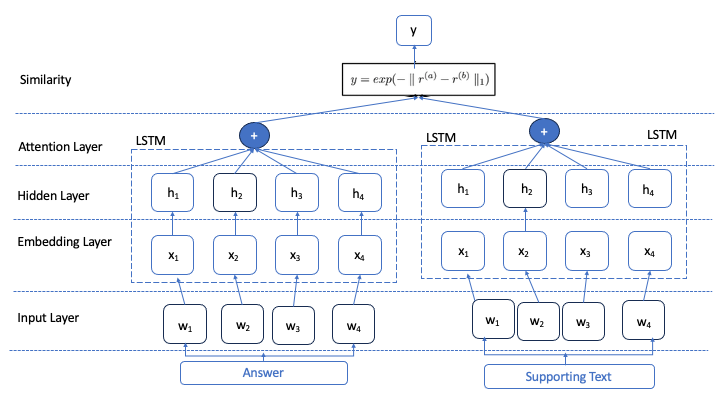}
  \caption{Siamese Networks for Text Similarity in Automated Answer Validation}
  \label{fig:siamese_arch}
\end{figure*}

A Siamese neural network \cite{bromley1993signature}, often referred to as a twin neural network, is a type of artificial neural network designed for comparing the similarity between two different input vectors by utilizing shared weights in parallel. One of the output vectors is typically precomputed, establishing a baseline for comparison against the other output vector. This concept can be likened to fingerprint comparison but is better described as a distance function used in locality-sensitive hashing. For the text similarity tast, the siamese network looks like the architecture in Figure \ref{fig:siamese_arch}.

Mathematically, the goal is to minimize the distance metric for similar objects while maximizing it for dissimilar ones. The most commonly used distance metric is the Euclidean distance, represented as:

\[
D(x, y) = \sqrt{\sum_{i=1}^{n}(x_i - y_i)^2}
\]

Here, \(x\) and \(y\) are vectors of dimension \(n\), and \(D(x, y)\) calculates the Euclidean distance between them.

\subsection{Learning in Siamese Networks}

Learning in Siamese networks can be achieved using various loss functions, with two common approaches being triplet loss and contrastive loss.

\subsubsection*{Triplet Loss}

In triplet loss, the network learns by comparing three inputs: an anchor sample, a positive sample, and a negative sample. The loss function encourages the network to minimize the distance between the anchor and positive samples while maximizing the distance between the anchor and negative samples. The triplet loss function can be defined as:

\[
L = \textit{max}(d(a, p) - d(a, n) + \textit{margin}, 0)
\]

Where:
\begin{itemize}
    \item \(L\) is the triplet loss.
    \item \(d(a, p)\) is the distance between the anchor and positive sample.
    \item \(d(a, n)\) is the distance between the anchor and negative sample.
    \item \(\textbf{margin}\) is a hyper parameter that enforces a minimum difference between positive and negative distances.
\end{itemize}

\subsubsection*{Contrastive Loss}

In contrastive loss, a weight decay or similar regularization technique is used in combination with the network to minimize the distance between similar objects and maximize the distance between dissimilar ones.

\section{Experiments}
\label{sec:experiments}

\subsection{Datasets}

The SciQ dataset \cite{Welbl2017CrowdsourcingMC} contains 13,679 crowdsourced science exam questions about Physics, Chemistry and Biology, among others. The questions are in multiple-choice format with 4 answer options each. For the majority of the questions, an additional paragraph with supporting evidence for the correct answer is provided.

\begin{itemize}
    \item \textbf{answers 1 - 4}: Candidate answers for the question (string).
    \item \textbf{correct\_answer}: The correct answer for the question (string).
    \item \textbf{support}: The supporting text for the question (string).
\end{itemize}

In addition to the above dataset, we are considering generating similar datasets in other domains using generative AI services like ChatGPT. Given the availability of question papers from competive exams, we might be able to generate correct answers and supporting text using ChatGPT. We can combine the question and the supporting text to validate the student answers.

\subsection{Benchmarks}
Benchmarks typically include datasets, metrics, evaluation protocols and baseline models. SciQ dataset by \cite{Welbl2017CrowdsourcingMC} is the obvious benchmark for this task. The dataset was generated in a crowdsourced approach. They employ a number of methods to generate multiple choice answers. They predominantly report accuracy as the metric. We propose to mostly follow their approach and report accuray of the Siamese network model in accurately selecting the correct answer among the multiple choices.

\subsection{Experimental setup}
We implemented our solutions using pytorch \cite{paszke2019pytorch} though we initially used keras for prototype models  For the application, we considered gradio, streamlit and fastAPI and chose streamlit. The deployment details are as discussed in Section \ref{sec:deployment}. We also used other packages and tools like nltk \cite{loper2002nltk} and pandas \cite{mckinney2011pandas}.

\subsection{SBERT}  

Our second baseline for automating answer validation uses SBERT (Sentence-BERT) model \cite{reimers2019sentence}, in conjunction with cosine similarity. It employs pre-trained transformer-based models to generate embeddings for text data, allowing for the comparison of question-answer pairs. The 'paraphrase-distilroberta-base-v1' variant of SentenceTransformer was chosen for this task. The embeddings were processed to remove question words and punctuation, ensuring data consistency. The resulting cosine similarity scores indicate the degree of similarity between text documents. This SBERT-based approach achieved a notable level of accuracy, demonstrating its effectiveness in selecting correct answers among multiple choices.

The SBERT and cosine similarity approach offers a practical means of capturing the underlying semantics of text data. It generates embeddings that encapsulate the context of questions and answer options, facilitating a deeper understanding of their relationships. The cosine similarity scores provide a straightforward measure of similarity. The reported accuracy rate reflects the approach's relevance in educational assessment systems and its potential to enhance the efficiency and accuracy of automated answer evaluation.

\subsection{Siamese Networks}

Siamese networks are a class of deep learning models designed for text similarity tasks. They play a significant role in assessing the similarity between two text sequences, making them valuable tools for various natural language processing applications. In the context of automated answer evaluation, Siamese networks prove effective in determining the resemblance between a question and its associated answer options. Their underlying architecture involves processing both input texts through shared layers, allowing the network to learn and extract essential features from the text data.

The Siamese network solution in this context is particularly focused on automating the process of selecting the correct answer from a set of options. To achieve this, it leverages the concept of textual similarity. During training, the network learns to distinguish between the correct answer and distractors by understanding the inherent connections between the question and each answer choice. This learning process relies on encoding text into numerical representations and computing similarity scores, which are then utilized to make informed decisions.

Compared to traditional methods like Bag of Words (BoW), Siamese networks offer a more nuanced approach to text similarity. While BoW represents text documents as high-dimensional vectors based on word frequencies, Siamese networks delve deeper by capturing semantic relationships between words and phrases. Additionally, Siamese networks achieve an accuracy rate of 79.60\% in this context, showcasing their effectiveness as a competitive solution for automated answer evaluation tasks when compared to baseline methods. Overall, Siamese networks provide a robust and data-driven approach to text similarity assessment, making them a valuable asset in the realm of educational assessment systems.

\subsection{Hyperparameter tuning}

Hyperparameter tuning is a critical step in enhancing machine learning model performance. In this study, we conducted a grid search to identify optimal hyperparameters for a Siamese network used for text similarity scoring in automated answer evaluation. The hyperparameters considered included learning rate, vocabulary size, hidden dimension size, and embedding dimension size. Each hyperparameter played a vital role in shaping the Siamese network's architecture and learning process.

We performed an exhaustive grid search, exploring a predefined range of values for each hyperparameter. The objective was to find the hyperparameter combination that maximized the network's accuracy in predicting text similarity.

We evaluated each hyperparameter configuration using a binary cross-entropy loss function and measured the model's accuracy on a held-out test dataset. The results highlighted the importance of selecting appropriate hyperparameters to improve the Siamese network's performance. Our findings emphasized the value of systematic hyperparameter tuning in optimizing the Siamese network's ability to discern text similarity effectively, particularly in automated answer evaluation tasks.

\subsection{Results}
\cite{Welbl2017CrowdsourcingMC} use accuracy as the primary metric for evaluation. But their work is for creating the dataset. Perplexity or blue score if the model generates the questions and answers. We report accuracy in predicting the correct answer. We can also use text similarity metrics like in \cite{metrics}. 

\begin{table}[!ht]
    \centering
    \begin{tabular}{lrr}
        \hline
        \textbf{Model} & \textbf{Dataset} & \textbf{Accuracy} \\
        \hline
        %Bag of Words & sciq & 84.20\% \\
        SBERT & sciq & 74.90\% \\
        Siamese Networks & sciq & 84.50\% \\
        \hline
    \end{tabular}
    \caption{Comparison of different text similarity solutions for automated answer validation}
    \label{tab:results}
\end{table}

The results of our experiments are as tabulated in Table \ref{tab:results}. Our Siamese Networks implementation significantly outperforms the SBERT Cosine Similarity solution. We believe this is significant result given both the models use a similarity function and we are using SBERT for comparing sentence similarity. 

\section{Discussion}

In this section, we'll discuss ablations, deployment considerations and future extensions of this work.

\subsection{Ablations}

We did ablations aimed at uncovering the nuances in our approach and determine which components contribute most significantly to the accuracy of our system.

\begin{figure}[!htb]
\begin{subfigure}{0.48\textwidth}
    \centering
    \includegraphics[width=\linewidth]{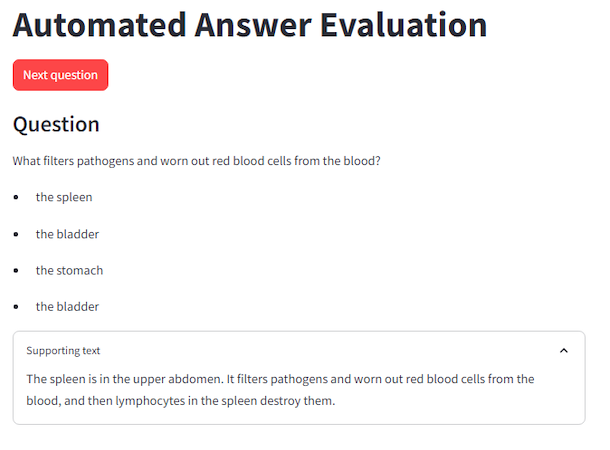}
    \caption{Questions and options}
    \label{fig:aav1}
\end{subfigure}
\begin{subfigure}{0.48\textwidth}
    \centering
    \includegraphics[width=\linewidth]{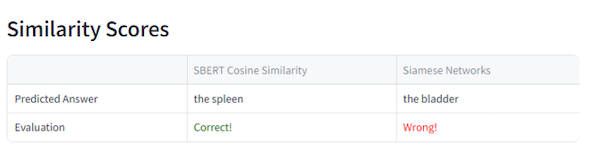}
    \caption{Similarity scores}
    \label{fig:aav2}
\end{subfigure}
\caption{Automated Answer Evaluation system deployed on streamlit}
\label{fig:aav}
\end{figure}

Firstly, we considered a simplified approach by exclusively relying on the provided answer options from the dataset. This meant discarding the original question entirely and evaluating answers solely based on their textual resemblance to the support text. While this approach demonstrated the feasibility of a straightforward answer matching mechanism, it lacked the contextual cues present in the question and often resulted in lower accuracy, as it missed the nuanced connections between the question and answers.

Next we experimented with a prefixing strategy, augmenting the answer options with the original question. This alteration expanded the length of the text being compared to the support text. By incorporating the question, we aimed to provide additional context and potentially enhance the model's ability to discern correct answers. This approach indeed yielded improvements in some cases, underlining the significance of contextual information in text similarity assessment.

Finally we extracted a portion of the support text to compare it directly with the answer options. This approach aimed to narrow down the scope of the comparison, concentrating on specific segments of the support text that were more relevant to the question. While this approach showed promise in certain scenarios, its effectiveness varied depending on the nature of the questions and the available support text segments.

These ablation experiments provided valuable insights into the mechanics of our answer evaluation system. They highlighted the importance of context, the potential trade-offs in text length, and the role of targeted text selection in achieving accurate results. Ultimately, these ablations allowed us to fine-tune our approach and make informed decisions about the most effective strategies for text similarity assessment in the context of automated answer evaluation.

\begin{table}[!ht]
    \centering
    \begin{tabular}{lr}
        \hline
        \textbf{Input Format} & \textbf{Accuracy} \\
        \hline
        Options alone & 84.50\% \\
        Options + question & 74.80\% \\
        Answer sentence selection & 25.05\% \\
        \hline
    \end{tabular}
    \caption{Siamese network with different input formats}
    \label{tab:ablations}
\end{table}

As shown in Table \ref{tab:ablations}, the siamese network works best when the options are compared with the full support text. The other variants we tried did not give any performance improvement.

%\begin{figure*}
%  \includegraphics[width=\textwidth]{aae.png}
%  \caption{Automated Answer Evaluation system deployed on streamlit}
%  \label{fig:aae}
%\end{figure*}

\subsection{Deployment}
\label{sec:deployment}
Our solution is deployed using Streamlit. We compared three frameworks Gradio \cite{gradio}, Streamlit \cite{streamlit} and FastAPI for our application deployment. We believe streamlit could be more suited to our needs. Streamlit helps to deploy the app in a public url without hosting charges. Gradio Spaces is another option but it is more suited towards LLM outputs. FastAPI will need our own hosting space and domains, which we wish to avoid at this point, though we may consider FastAPI later in case we decide to scale the solution later. As shown in Figure \ref{fig:aav}, we display one question from the test set and evaluate the multiple choices using different models. We also accept user answers in a free-form text area.

\subsection{Extensions}
\label{sec:extension}

In this section, we'll discuss couple of applications where automated answer validation can be used. While our implementation uses a DistilBERT model, this work can be implemented with most of the large language models available in the literature. Likewise the similarity model can be Siamese Network or any comparable similarity model.

\subsection*{Automated Answer Evaluation}
Building upon the existing automated answer validation system, an important extension involves refining the evaluation process. This can be achieved by incorporating advanced natural language processing techniques to not only check for textual similarity but also assess the semantic correctness of answers. For instance, the system can leverage semantic role labeling and dependency parsing to understand the relationships between words in a sentence. Additionally, sentiment analysis and opinion mining could be integrated to evaluate whether an answer is biased or subjective. These enhancements would lead to a more comprehensive and nuanced assessment of answers, providing valuable insights beyond mere text matching.

%\begin{figure}
%  \includegraphics[width=\columnwidth]{figures/aav3.png}
%  \caption{Users can directly enter answers for evaluation}
%  \label{fig:aav3}
%\end{figure}

%\subsection{Datascience Questions}
%Expanding the scope of the solution to encompass data science questions is another promising avenue. Data science questions often require a blend of domain-specific knowledge and technical expertise. To address this, the system can be extended to incorporate specialized data science models and libraries. It can leverage pre-trained models for tasks like named entity recognition on data-related terms, allowing it to better understand and evaluate answers pertaining to data manipulation, analysis, and visualization. Additionally, the system can integrate with data science libraries like Pandas and NumPy to validate code-based answers effectively. This extension would cater to the growing demand for automated evaluation in the data science and analytics domain.

\subsection*{Answering with LLMs}
Leveraging Large Language Models (LLMs) like GPT-3 and BERT is a promising direction for enhancing the answer validation system. By incorporating these models, the system can not only assess answer quality but also generate high-quality responses to questions. LLMs have the ability to understand context, generate coherent explanations, and provide detailed responses. They can be fine-tuned on specific question types and domains to ensure accurate and domain-aware answers. Additionally, LLMs can assist in handling ambiguous questions by generating clarifying queries or requesting additional information when necessary. This extension empowers the system to not only validate answers but also actively participate in answering questions, making it a more interactive and intelligent tool for users.

\section*{Conclusion}
We have implemented a solution for automated answer validation for question answering. By using text similarity models, we are able to validate the user given answer provided we have an answer key. This solution could be used to automate the evaluation of answers in educational institutions. We also discussed extensions to this work like using large language models to directly answer the question given the supporting text.

\section*{Acknowledgement}
This work was done as part of the capstone project in the Advanced Data Science program offered by the Department of Computational and Data Sciences in the Indian Institute of Science, Bengaluru and Talent Sprint. We thank Shashikumaar Ganesan, Deepak Subramani, and Surobhi Lahiri for their valuable feedback. We thank Srihai N for all the discussions and suggestions throughout the implementation. We thank Chandhini Ramesh and Abhishek Verma for their encouragement.

%\clearpage
\bibliography{references}

\begin{thebibliography}{21}
\expandafter\ifx\csname natexlab\endcsname\relax\def\natexlab#1{#1}\fi

\bibitem[{Abid et~al.(2023)Abid, Abdalla, Wexler, Zhang, Meng, and Peng}]{gradio}
Abubakar Abid, Ali Abdalla, James Wexler, Michael Zhang, Amanda Meng, and Andi Peng. 2023.
\newblock Gradio.
\newblock \url{https://www.gradio.app/}.
\newblock Accessed: Dec 07, 2023.

\bibitem[{Bromley et~al.(1993)Bromley, Guyon, LeCun, S{\"a}ckinger, and Shah}]{bromley1993signature}
Jane Bromley, Isabelle Guyon, Yann LeCun, Eduard S{\"a}ckinger, and Roopak Shah. 1993.
\newblock Signature verification using a" siamese" time delay neural network.
\newblock \emph{Advances in neural information processing systems}.

\bibitem[{Clark et~al.(2016)Clark, Etzioni, Khot, Sabharwal, Tafjord, Turney, and Khashabi}]{clark2016combining}
Peter Clark, Oren Etzioni, Tushar Khot, Ashish Sabharwal, Oyvind Tafjord, Peter Turney, and Daniel Khashabi. 2016.
\newblock Combining retrieval, statistics, and inference to answer elementary science questions.
\newblock In \emph{Proceedings of the AAAI Conference on Artificial Intelligence}, volume~30.

\bibitem[{Clark et~al.(2013)Clark, Harrison, and Balasubramanian}]{clark2013study}
Peter Clark, Philip Harrison, and Niranjan Balasubramanian. 2013.
\newblock A study of the knowledge base requirements for passing an elementary science test.
\newblock In \emph{Proceedings of the 2013 workshop on Automated knowledge base construction}, pages 37--42.

\bibitem[{Gollapalli and Ng(2022)}]{gollapalli2022qsts}
Sujatha~Das Gollapalli and See~Kiong Ng. 2022.
\newblock Qsts: A question-sensitive text similarity measure for question generation.
\newblock In \emph{Proceedings of the 29th International Conference on Computational Linguistics}, pages 3835--3846.

\bibitem[{Gupta(2018)}]{metrics}
Sanket Gupta. 2018.
\newblock \href {https://towardsdatascience.com/overview-of-text-similarity-metrics-3397c4601f50} {Overview of text similarity metrics in python}.

\bibitem[{Khashabi et~al.(2016)Khashabi, Khot, Sabharwal, Clark, Etzioni, and Roth}]{khashabi2016question}
Daniel Khashabi, Tushar Khot, Ashish Sabharwal, Peter Clark, Oren Etzioni, and Dan Roth. 2016.
\newblock Question answering via integer programming over semi-structured knowledge.
\newblock \emph{arXiv preprint arXiv:1604.06076}.

\bibitem[{Khot et~al.(2015)Khot, Balasubramanian, Gribkoff, Sabharwal, Clark, and Etzioni}]{khot2015exploring}
Tushar Khot, Niranjan Balasubramanian, Eric Gribkoff, Ashish Sabharwal, Peter Clark, and Oren Etzioni. 2015.
\newblock Exploring markov logic networks for question answering.
\newblock In \emph{Proceedings of the 2015 conference on empirical methods in natural language processing}, pages 685--694.

\bibitem[{Li et~al.(2023)Li, Shen, Sun, Zhang, Nan, Gao et~al.}]{li2023using}
Meiting Li, Xifeng Shen, Yuanyuan Sun, Weining Zhang, Jiale Nan, Dongping Gao, et~al. 2023.
\newblock Using semantic text similarity calculation for question matching in a rheumatoid arthritis question-answering system.
\newblock \emph{Quantitative Imaging in Medicine and Surgery}, 13(4):2183.

\bibitem[{Li and Clark(2015)}]{li2015answering}
Yang Li and Peter Clark. 2015.
\newblock Answering elementary science questions by constructing coherent scenes using background knowledge.
\newblock In \emph{Proceedings of the 2015 Conference on Empirical Methods in Natural Language Processing}, pages 2007--2012.

\bibitem[{Loper and Bird(2002)}]{loper2002nltk}
Edward Loper and Steven Bird. 2002.
\newblock Nltk: The natural language toolkit.
\newblock \emph{arXiv preprint cs/0205028}.

\bibitem[{Malkiel et~al.(2022)Malkiel, Ginzburg, Barkan, Caciularu, Weill, and Koenigstein}]{malkiel2022interpreting}
Itzik Malkiel, Dvir Ginzburg, Oren Barkan, Avi Caciularu, Jonathan Weill, and Noam Koenigstein. 2022.
\newblock Interpreting bert-based text similarity via activation and saliency maps.
\newblock In \emph{Proceedings of the ACM Web Conference 2022}, pages 3259--3268.

\bibitem[{McKinney et~al.(2011)}]{mckinney2011pandas}
Wes McKinney et~al. 2011.
\newblock pandas: a foundational python library for data analysis and statistics.
\newblock \emph{Python for high performance and scientific computing}, 14(9):1--9.

\bibitem[{Paszke et~al.(2019)Paszke, Gross, Massa, Lerer, Bradbury, Chanan, Killeen, Lin, Gimelshein, Antiga et~al.}]{paszke2019pytorch}
Adam Paszke, Sam Gross, Francisco Massa, Adam Lerer, James Bradbury, Gregory Chanan, Trevor Killeen, Zeming Lin, Natalia Gimelshein, Luca Antiga, et~al. 2019.
\newblock Pytorch: An imperative style, high-performance deep learning library.
\newblock \emph{Advances in neural information processing systems}, 32.

\bibitem[{Reimers and Gurevych(2019)}]{reimers2019sentence}
Nils Reimers and Iryna Gurevych. 2019.
\newblock Sentence-bert: Sentence embeddings using siamese bert-networks.
\newblock \emph{arXiv preprint arXiv:1908.10084}.

\bibitem[{Sachan et~al.(2016)Sachan, Dubey, and Xing}]{sachan2016science}
Mrinmaya Sachan, Avinava Dubey, and Eric~P Xing. 2016.
\newblock Science question answering using instructional materials.
\newblock \emph{arXiv preprint arXiv:1602.04375}.

\bibitem[{Singh(2019)}]{singh2019siamese}
Prabhnoor Singh. 2019.
\newblock \href {https://medium.com/@prabhnoor0212/siamese-network-keras-31a3a8f37d04} {Siamese network keras for image and text similarity}.

\bibitem[{Team(2023)}]{streamlit}
Streamlit Team. 2023.
\newblock Streamlit.
\newblock \url{https://www.streamlit.io/}.
\newblock Accessed: Dec 07, 2023.

\bibitem[{Vallejo et~al.(2022)Vallejo, Baldwin, and Frermann}]{vallejo2022evaluating}
Gisela Vallejo, Timothy Baldwin, and Lea Frermann. 2022.
\newblock Evaluating the examiner: The perils of pearson correlation for validating text similarity metrics.
\newblock In \emph{Proceedings of the The 20th Annual Workshop of the Australasian Language Technology Association}, pages 130--138.

\bibitem[{Viji and Revathy(2022)}]{viji2022hybrid}
D~Viji and S~Revathy. 2022.
\newblock A hybrid approach of weighted fine-tuned bert extraction with deep siamese bi--lstm model for semantic text similarity identification.
\newblock \emph{Multimedia Tools and Applications}, 81(5):6131--6157.

\bibitem[{Welbl et~al.(2017)Welbl, Liu, and Gardner}]{Welbl2017CrowdsourcingMC}
Johannes Welbl, Nelson~F. Liu, and Matt Gardner. 2017.
\newblock \href {https://api.semanticscholar.org/CorpusID:1553193} {Crowdsourcing multiple choice science questions}.
\newblock \emph{ArXiv}, abs/1707.06209.

\end{thebibliography}
%\bibliographystyle{acl_natbib}

%\appendix

%\section{Example Appendix}
%\label{sec:appendix}

%This is an appendix.

\end{document}